\pdfoutput=1

\documentclass[11pt]{article}

\usepackage{acl}

\usepackage{times}
\usepackage{latexsym}

\usepackage{graphicx}
\usepackage{amsmath,amsfonts,amssymb}
\usepackage{mathtools}
\usepackage{caption}
\usepackage{lipsum}
\usepackage{booktabs}
\usepackage{xcolor}
\usepackage{soul}

\usepackage{color, colortbl}
\definecolor{Gray}{gray}{0.9}
\definecolor{ao(english)}{rgb}{0.0, 0.5, 0.0}
\definecolor{cardinal}{rgb}{0.77, 0.12, 0.23}
\usepackage{url}
\usepackage{float}
\usepackage{tabularx}
\usepackage{appendix}
\usepackage{placeins}
\usepackage{bm}
\makeatletter
\newcommand{\ssymbol}[1]{^{\@fnsymbol{#1}}}

\usepackage[T1]{fontenc}

\usepackage[utf8]{inputenc}

\usepackage{microtype}

\title{Improving Topic Segmentation by Injecting Discourse Dependencies}

\author{First Author \\
  Affiliation / Address line 1 \\
  Affiliation / Address line 2 \\
  Affiliation / Address line 3 \\
  \texttt{email@domain} \\\And
  Second Author \\
  Affiliation / Address line 1 \\
  Affiliation / Address line 2 \\
  Affiliation / Address line 3 \\
  \texttt{email@domain} \\}
  
\author{Linzi Xing , Patrick Huber , Giuseppe Carenini\\
  Department of Computer Science \\
  University of British Columbia \\
  Vancouver, BC, Canada, V6T 1Z4 \\ 
  {\tt \{lzxing, huberpat, carenini\}@cs.ubc.ca}}

\date{}

\begin{document}
\maketitle

\begin{abstract}
Recent neural supervised topic segmentation models achieve distinguished superior effectiveness over unsupervised methods, with the availability of large-scale training
corpora sampled from \textit{Wikipedia}. These models may, however, suffer from limited robustness and transferability caused by exploiting simple linguistic cues for prediction, but overlooking more important inter-sentential topical consistency. To address this issue,  we present a discourse-aware neural topic segmentation model with the injection of above-sentence discourse dependency structures to encourage the model make topic boundary prediction based more on the topical consistency between sentences. Our empirical study on English evaluation datasets shows that injecting above-sentence discourse structures 
to a neural topic segmenter with our proposed strategy can substantially improve its performances on intra-domain and out-of-domain data,
with little increase of model's complexity.

\end{abstract}

\section{Introduction}

Topic segmentation is a fundamental NLP task with the goal to separate textual documents 
into coherent segments (consisting of one or more sentences), following the document's underlying topical structure.
The structural knowledge obtained from topic segmentation has been shown to play a vital role in key NLP downstream tasks, such as document summarization \cite{mitra-etal-1997-automatic, riedl-biemann-2012-text, xiao-carenini-2019-extractive}, question answering \cite{oh07, dennis-2017-core} and dialogue modeling \cite{topic_dialogue_2020, ijcai2020-517}.
The aim of topic segmentation makes it tightly connected to related research areas aiming to understand the latent structure of long and potentially complex text.
\begin{figure}
    \centering
    \includegraphics[width=3.0in]{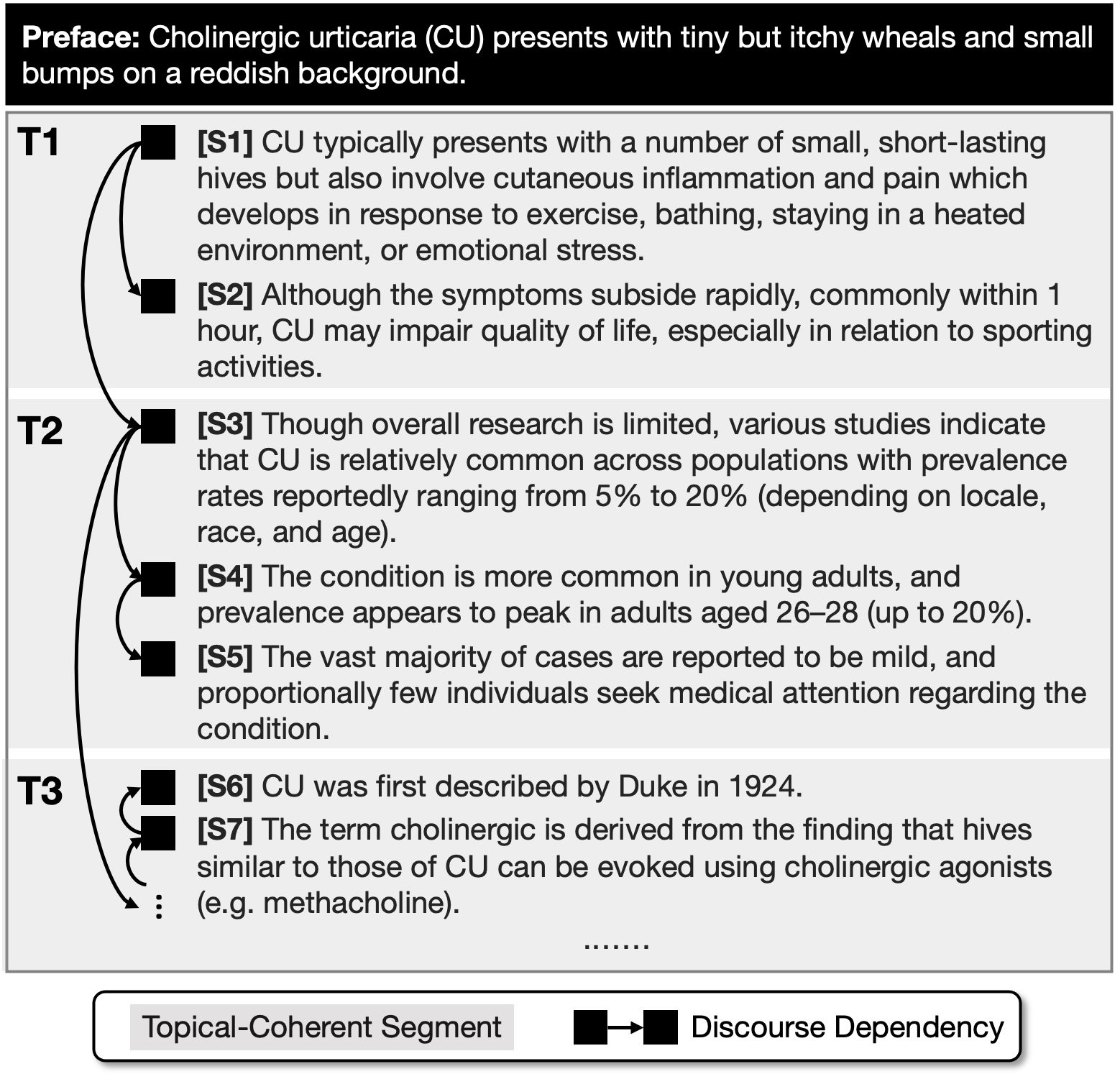}
    \caption{An example article about Cholinergic Urticaria (CU) sampled from the \textit{en\_disease} portion of Wiki-Section dataset \cite{arnold-etal-2019-sector}. 
    Left: discourse dependency structure predicted by the Sent-First discourse parser \cite{zhou-feng-2022-improve}.}
    \label{fig:wiki_example}
\end{figure}
Specifically, understanding the semantic and pragmatic underpinnings of a document can arguably support the task of separating continuous text into topical segments. To this end, discourse analysis and discourse parsing provide the means to understand and infer the semantic and pragmatic relationships underlying complete documents, well aligned with the local text coherence and highly correlated to the inter-sentential topical consistency, as shown in \citet{louis-nenkova-2012-coherence} and \citet{muangkammuen-etal-2020-neural}. With a variety of linguistic theories proposed in the past, such as the Rhetorical Structure Theory (RST) \cite{mann1988rhetorical}, the lexicalized discourse framework \cite{webber-etal-2003-anaphora} (underlying PDTB), and the Segmented Discourse Representation Theory (SDRT) \cite{asher1993reference, asher2003logics}, we follow the RST framework in this work (1) as we focus on monologue text (as compared to dialogue frameworks, such as SDRT) and (2) since RST postulates complete discourse trees spanning whole documents, directly aligned with the topical structure of complete documents \cite{huber2021predicting}.

We further motivate the synergistic relationship between topic segmentation and discourse analysis/parsing in Figure~\ref{fig:wiki_example}, showing anecdotal evidence of the alignment between the document's topical structure and the respective RST-style discourse dependency graph. Starting from a sequence of sentences, the task of topic segmentation addresses the problem of splitting the given \textit{Wikipedia} article into an ordered set of topical-coherent fragments (here: \texttt{T1}, \texttt{T2} and \texttt{T3}) by predicting topical boundaries. As shown in the example, the document discourse tree is indicative of the topical structure of the document, as discourse dependencies occur considerably more often within a topic segment than across topic segments.


Given significant influence on a variety of real-world tasks, topic segmentation is an active research area in the field of NLP. As such, modern, neural methods for monologue topic segmentation are proposed by formulating the task as a sentence-level sequence labeling problem, trained and evaluated on the large-scale \textit{Wikipedia} dataset \cite{xing-etal-2020-improving, glavas-2020-two, barrow-etal-2020-joint, lo-etal-2021-transformer-pre}. These Wikipedia articles are well-suited for the task of topic segmentation, providing natural section marks which can be reasonably used as ground-truth segment boundaries \cite{koshorek-etal-2018-text, arnold-etal-2019-sector}, superseding previously proposed 
unsupervised methods \cite{hearst-1997-text, galley-etal-2003-discourse, eisenstein-barzilay-2008-bayesian, Song2016DialogueSS}. Despite the significant improvements achieved by neural supervised topic segmentation models, 
it remains unclear if these topic segmenters effectively learn to cluster sentences into topical-coherent pieces based on 
the (document-level) topical consistency, or solely exploit superficial patterns (e.g., simple linguistic cues)  
in the training domain.



To address this challenge, in this paper,
we propose a more discourse-aware neural topic segmentation model. We thereby inject above-sentence discourse structures into basic topic segmenter to encourage the model to base its topic boundary prediction more explicitly on the topical consistency between sentences.
More specifically, we propose to exploit a discourse dependency parser pre-trained on out-of-domain data to induce inter-sentential discourse dependency trees. Subsequently, we convert the dependency tree into a directed discourse graph with sentences as nodes and discourse dependencies as edges. With the generated discourse graph, a Graph Attention Network (GAT) \cite{velickovic2018graph} is used to encode sentences as discourse-contextualized representations by aggregating information from neighboring sentence nodes in the graph. 
Finally, the discourse-infused sentence representations are concatenated with standard encodings for segment boundary prediction.

In our empirical study conducted on English evaluation datasets, we show that: ($i$) Injecting discourse structures can substantially improve the performance of the basic neural topic segmentation model on three datasets. ($ii$) Our novel, discourse-enhanced topic segmenter is more robust compared to the basic neural model in settings that require domain transfer, showing superior performance on four challenging real-world test sets, to confirm the improved domain-independence. ($iii$) Even if our proposal has inferior accuracy against a state-of-the-art segmenter sharing the same basic architecture, it does achieve significantly better efficiency assessed by model's parameter size and speeds for learning and inference, which makes it potentially more favorable in real-world use.

\section{Related Work}
\paragraph{Topic Segmentation} aims to reveal important aspects of the semantic structure of a document by splitting a sequence of sentences into topic-coherent textual units. Typically, computational topic segmentation models can be broadly separated into supervised and unsupervised approaches. Early topic segmentation methods usually fall into the category of unsupervised approaches, mainly due to the prevalent data sparsity issue at the time. Based on predicting the coherence between sentences through shallow (surface-level) features, unsupervised models reach a limited understanding of the contextualized structure of documents by merely relying on easy-to-extract but barely effective features for the similarity measurement between sentences (i.e., the degree of token overlap between two sentences) \cite{hearst-1997-text, eisenstein-barzilay-2008-bayesian}.
Improving on the unsupervised topic segmentation paradigm, researchers started to address this issue by introducing pre-trained neural language models (LMs), trained on massive dataset \cite{topic_dialogue_2020, solbiati2021unsupervised, xing-carenini-2021-improving}.
Some works show that the signal captured in pre-trained LMs (e.g., BERT \cite{devlin2019bert}) 
are more indicative of topic relevance between sentences than early surface-level features. However, these proposed strategies of integrating BERT into the topic segmentation framework solely exploit BERT to induce 
dense encodings and further 
compute reciprocal sentence similarities.
While this constitues a reasonable first step, the considerable gap between the training objective of LMs and topic segmentation task requires further efforts 
along this line of work \cite{sun-etal-2022-sentence}. 

More recently, the data sparsity issue has been alleviated by the proposal of large-scale corpora sampled from \textit{Wikipedia} (e.g., Wiki-727k \cite{koshorek-etal-2018-text} and Wiki-Section \cite{arnold-etal-2019-sector}), in which well-structured articles with their section marks are used as gold labels for segment boundaries. As a result, neural supervised topic segmenters started to gain attention by reaching greater effectiveness and efficiency compared to previously proposed unsupervised approaches. These supervised topic segmenters typically follow a common strategy which formulates the task as a sentence-level sequence labeling problem. More specifically, by assigning binary labels to each sentence, models infer the likelihood of a sentence to be a topic segment boundary \cite{ koshorek-etal-2018-text, arnold-etal-2019-sector, barrow-etal-2020-joint, lo-etal-2021-transformer-pre}. However, we believe that current models, besides reaching promising performance, potentially favour simple linguistic cues over effective measurements for semantic cohesion, restricting their application to narrow domains.
Some recent works have attempted to address this limitation via explicitly integrating coherence modeling components into segmenters
\cite{xing-etal-2020-improving, glavas-2020-two}. However, compared to our objective in this work, these proposed coherence modeling strategies are either (i) only taking two adjacent sentences into account, limiting the additional module to extremely local contexts, or (ii) discriminating real documents from artificially ``incoherent" texts, resulting in implicit and synthetic negative training samples and heavy parameter size caused by modeling multiple tasks simultaneously.  

In contrast, we propose an effective method to integrate the document discourse (dependency) structure into neural topic segmentation frameworks, following the intuition that above-sentence discourse structure are indicative of text coherence and topical consistency, providing a more global and interpretable source of information for better
topic transition prediction.
 
\paragraph{Discourse Analysis and Parsing} analyze and generalize the underlying semantic and pragmatic structure of a coherence document (called a discourse). As an important upstream task in the field of NLP, discourse analysis 
proposes elaborate frameworks and theories to describe the textual organization of a document. To this end, a variety of popular discourse theories proposed in the past, such as (besides others) the Rhetorical Structure Theory (RST) \cite{mann1988rhetorical} and the lexicalized discourse framework \cite{webber2003anaphora} for monologues as well as the Segmented Discourse Representation Theory (SDRT) \cite{asher1993reference, asher2003logics} for dialogues. 
Among these theories, the RST discourse theory postulates a single, complete discourse tree for monologue documents, while the lexicalized discourse framework only focuses on local discourse connectives within and between adjacent sentences. Focusing on the connection between discourse information and topic segmentation, we employ the RST discourse theory in this work, most aligned with the requirement to capture topical coherence.


Building on human annotated discourse treebanks, a mix of traditional and neural discourse parsers have been proposed over the last decades, with traditional approaches mainly exploiting surface-level features through Support-Vector Machines (SVMs) \cite{hernault2010hilda, ji-eisenstein-2014-representation, wang-etal-2017-two} or Conditional Random Fields (CRFs) \cite{joty-etal-2015-codra, feng2014linear}. 
On the other hand, neural models achieve similar or superior results on RST discourse parsing, with models using either custom architectures \cite{yu-etal-2018-transition, liu2018learning} or pre-trained LMs (e.g. BERT \cite{zhou-feng-2022-improve}, RoBERTa \cite{guz-etal-2020-unleashing}, SpanBERT \cite{guz-carenini-2020-coreference}).
\begin{figure*}
    \centering
    \includegraphics[width=6.3in]{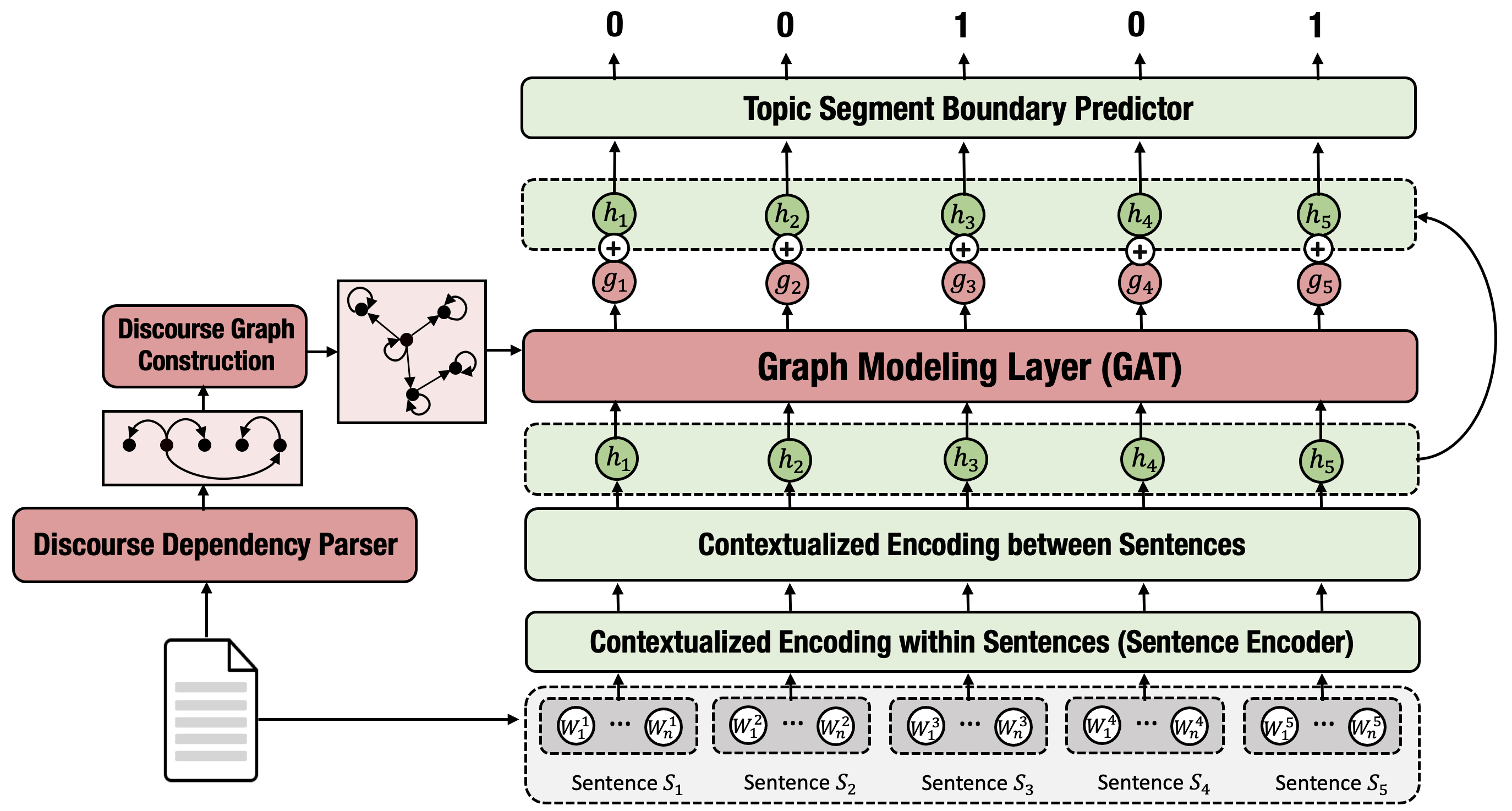}
    \caption{The overall architecture of our discourse-infused topic segmentation model.
    }
    \label{fig:proposed_model}
\end{figure*}
In this work, we generate discourse dependency trees from a BERT-based neural dependency parser proposed in \citet{zhou-feng-2022-improve}, since: (i) The parser follows the intuition that information, and hence structures, in sentences are oftentimes ``self-contained''. Therefore, 
it predicts the interactions between EDUs of the same sentence in a first stage and subsequently predicts the inter-sentential discourse structures, which aligns well with our objective of sentence-level topic segmenation. (ii) 
The parser by \citet{zhou-feng-2022-improve} makes direct prediction of dependency discourse structures, alleviating the potential error caused by converting constituency structures into their respective dependency trees.

\section{Methodology}
As shown in Figure~\ref{fig:proposed_model}, our proposed discourse-aware neural topic segmentation model comprises two components: the \textit{Hierarchical Topic Segmenter} and \textit{Discourse Graph Modeling}, highlighted in green and red respectively. Discourse Graph Modeling further comprises of a \textit{Discourse Graph Construction} and \textit{Graph Modeling} component.

\subsection{Basic Model: Hierarchical Topic Segmenter} 
\label{sec:basic_model}
The basic architecture of our proposal is adopted from the basic model in
\citet{xing-etal-2020-improving}, consisting of two hierarchical layers: First, a sentence encoder contextualizes individual sentences, followed by the second layer, conditioning sentences on the complete document. Following the settings in \citet{xing-etal-2020-improving}, we adopt the attention BiLSTM architecture\footnote{We also considered Transformer as the backbone of contextualized encoder, but eventually chose BiLSTM for its superior performance.} 
for each layer and enhance the encodings with pre-trained BERT embeddings. Formally, given a document $D$ as a sequence of $n$ sentences, the sentence encoder (bottom component in Figure~\ref{fig:proposed_model}) yields the embedding for each individual sentence. Based on the obtained encodings, the document-level contextualization layer returns an ordered set of hidden states $\bm{H} = \{ \bm{h}_1, ..., \bm{h}_{n} \}$. Next, a simple multilayer perceptron (MLP) with a final softmax activation serves as a binary topic boundary predictor based on a threshold $\tau$, tuned on the validation set. During training, we optimize the model in accordance to the cross-entropy loss, while at inference time, every sentence (except the last sentence\footnote{We remove the last sentence from the sequence for prediction since it is per definition the end of the last segment.}) with a probability  $\geq \tau$ is considered as the end of a segment.

\subsection{Discourse Graph Modeling}
Our goal is to inject inter-sentential discourse dependency structures into the task of topic segmentation. We believe that the additional, structural information is thereby well aligned with the topical consistency between sentences, hence, suited to guide the prediction of topic transitions. To integrate the discourse information into the
basic model described in section~\ref{sec:basic_model}, we first generate an above-sentence discourse dependency tree $\bm{T}_D$ for the document. Specifically, we utilize the discourse dependency parsing model proposed in \citet{zhou-feng-2022-improve}, reaching
state-of-the-art performance for discourse tree construction and relation type identification in multiple language settings. The ``Sent-First'' parser \cite{zhou-feng-2022-improve} further fits the aim of our proposal due to its two-staged approach, first generating discourse trees within sentences and subsequently combining sentence-level sub-trees. This hard constraint allows us to exclusively obtain above-sentence discourse structures, avoiding potentially leaky sub-trees \cite{joty-etal-2015-codra}. 
Regarding the discourse relations attached to every head-dependent pair (discourse dependency), we follow the observation in \citet{xu-etal-2020-discourse}, stating that the agreement between the type of rhetorical relation is usually lower and more ambiguous, 
to leave them for future work to avoid error propagation.

In contrast to the original proposal in \citet{zhou-feng-2022-improve}, training and testing their dependency discourse parser on one corpus (i.e., SciDTB
\cite{yang-li-2018-scidtb}), we believe that a mixture of several diverse and publicly available discourse treebanks with different document lengths and domains can increase the parser's robustness on new and unseen genres. Therefore, we retrain the parser on a mixture of RST-DT\footnote{\url{catalog.ldc.upenn.edu/LDC2002T07}} 
\cite{carlson2002rst}, GUM\footnote{\url{corpling.uis.georgetown.edu/gum}} 
\cite{Zeldes2017}, SciDTB\footnote{ \url{https://github.com/PKUTANGENT/SciDTB}} 
\cite{yang-li-2018-scidtb} and COVID19-DTB\footnote{\url{https://github.com/norikinishida/biomedical-discourse-treebanks}} 
\cite{10.1162/tacl_a_00451}.
More specifically, we combine those discourse treebanks and randomly split the aggregated corpus into 80\% training,
10\% validation, 10\% test data. The parser retrained on our combined training portion achieves an Unlabeled Attachment Score (UAS) of 58.6 on the test portion. We show additional key dataset statistics for each treebank used in this paper in Table~\ref{tab:treebanks}.

\begin{table}
\centering
\scalebox{0.9}{
\begin{tabular}{l | @{\space\space\space} c@{\space\space\space} c@{\space\space\space} c@{\space\space\space}}

\specialrule{.1em}{.05em}{.05em}
\textbf{Treebank} & \# of doc & \# sent/doc & \# edu/doc  \\
\hline
 RST-DT & 385 & 22.5 & 56.6  \\
 GUM & 150 & 49.3 & 114.2 \\
 SciDTB &  1,355 & 5.3 & 14.1 \\
 COVID19-DTB & 300 & 7.8 & 20.0 \\
\specialrule{.1em}{.05em}{.05em}
\end{tabular}
}
\caption{\label{tab:treebanks} Key dataset statistics of the discourse treebanks used for retraining the Sent-First discourse parser \cite{zhou-feng-2022-improve}.
}
\end{table}

After training the discourse parser to infer a discourse dependency tree $\bm{T}_{D}$ for document $D$, we convert the tree structure into a discourse graph $\bm{G}_{D}$ (as a binary matrix).
Formally, we initialize the graph $\bm{G}_{D}$ as a $n \times n$ identity 
matrix $\bm{G}_D = \bm{I}_{n, n} $, connecting every node to itself. Afterwards, we
fill in the remaining cells by assigning $\bm{G}_{D}[i][j] = 1$ iff $ \exists~ \bm{T}_{D}(i \rightarrow j)$, with $i$, $j$ indexing the head and dependant sentences in the document, respectively. 
Using the binary matrix representation of $\bm{G}_{D}$, we apply the multi-layer Graph Attention Network (GAT) \cite{velickovic2018graph} to update sentence encodings following 
the discourse graph. More specifically, with the discourse graph matrix $\bm{G}_{D}$ and the contextualized representations $\bm{H} = \{ \bm{h}_1, ..., \bm{h}_{n} \}$ described in section~\ref{sec:basic_model}, within each graph attentional layer, we perform self-attention on the sentence nodes. Taking the $l$th layer as an example, 
we compute the attention coefficient $\alpha_{i,j}$ between sentence nodes $i$, $j$ as:
\vspace{-0.5ex}
\begin{equation}
    \alpha^{l}_{ij} = softmax(e^{l}_{ij}) = \frac{exp(e^l_{ij})}{\sum_{k \in \mathcal{N}_{i}} exp(e^{l}_{ik})}, \label{eq:1}
\end{equation}
\vspace{-1.5ex}
\begin{equation}
    e^{l}_{ij} = LeakyReLU(\bm{a}_{l}^{T}[\bm{W}_{l}\bm{g}^l_i||\bm{W}_{l}\bm{g}^l_j]) \label{eq:2}
\end{equation}
where $\bm{W}_l$ and $\bm{a}_l$ are learnable parameters for layer $l$ and $^{T}$ is the transposition operation. $\mathcal{N}_i$ denotes the direct neighborhood of node $i$ in the graph ($G_D[i][\cdot] = 1$). As the node representation input of the first GAT layer ($l = 0$), $\bm{g}^{0}_{i} = \bm{h}_{i} \in H$. Once attention coefficients are obtained, we compute the intermediate node representation $\bm{z}_{i}^{l}$ for sentence node $i$ at layer $l$ by aggregating information from neighboring nodes as:
\begin{equation}
    \bm{z}^{l}_{i} = \sum_{j \in \mathcal{N}_{i}}\alpha^{l}_{ij}\bm{W}_{l}\bm{g}^l_j \label{eq:3}
\end{equation}
Following the step in \citet{huang-etal-2020-grade}, we combine the intermediate node representation $\bm{z}_{i}^{l}$ with the input of this layer $\bm{g}_{i}^{l}$ to get the updated node representation $\bm{g}_{i}^{l+1}$ as the input for the next layer:
\begin{equation}
    \bm{g}^{l+1}_{i} = ELU(\bm{g}^l_i + \bm{z}_{i}^{l}) \label{eq:4}
\end{equation}
where ELU denotes an exponential linear unit \cite{Clevert2016FastAA}. With the output $\bm{g}_i$ from the last layer of GAT, we concatenate it together with $\bm{h}_i$ and further feed $[\bm{h}_i ; \bm{g}_{i}]$ into the predictor layer for segment boundary prediction.

\section{Experiments}
In order to quantitatively evaluate the effectiveness, generality and efficiency of our proposal, we conduct three sets of experiments to compare our topic segmentation approach against a variety of baselines and previous models. Namely, we assess the performance of our model in regards to the \textit{Intra-Domain Segment Inference Performance}, \textit{Domain Transfer Segment Inference Performance}, and conduct an additional \textit{Efficiency Analysis}.

\subsection{Datasets}

\subsubsection{Intra-Domain Datasets}
For the set of intra-domain segment inference experiments, we train and test models within the same domain (here: on the same corpus). We thereby choose three diverse corpora (see Table~\ref{tab:stats_intra} for more details) for the intra-domain evaluation:
\paragraph{Choi \cite{choi-2000-advances}.} This 
corpus consists of 920 articles artificially generated by randomly combining passages from the Brown corpus. The datapoints in this dataset are not human written, leading us to solely use this corpus for a preliminary performance assessment for topic segmentation models in a 80\% (train)/10\%(dev)/10\%(test) data-split. 

\begin{table}
\centering
\scalebox{0.93}{
\begin{tabular}{l | @{\space\space\space} c@{\space\space\space\space\space} c@{\space\space\space\space\space} c@{\space\space\space\space} }

\specialrule{.1em}{.05em}{.05em}
\textbf{Dataset} & \# of doc & \# sent/seg & \# seg/doc  \\
\hline
 CHOI & 920 & 7.4 & 10.0  \\
 RULES & 4,461 & 7.4 & 16.0 \\
 SECTION & 21,376 & 7.2 & 7.9 \\
\specialrule{.1em}{.05em}{.05em}
\end{tabular}
}
\caption{\label{tab:stats_intra} Statistics of the datasets used in intra-domain experiments.}
\end{table}

\paragraph{Rules \cite{bertrand2018hall}.} This corpus consists of 4,461 documents about regulation discussion published in the Federal Register\footnote{\url{https://www.govinfo.gov/}} by U.S. federal agencies. 
Since each paragraph is about 
one particular regulation and all regulations covered by one document are under the same category, we deem it as a reasonably coherent data source for topic segmentation evaluation with the paragraph breaks as ground-truth segment boundaries. We split this dataset into training, validation and test sets with the default 80\%, 10\%, 10\% data-split.

\paragraph{Wiki-Section (Section) \cite{arnold-etal-2019-sector}.} This corpus originally contains Wikipedia articles in both English and German. The English portion of the dataset, which we use for our intra-domain experiment, consists of around 3.6k articles about diseases and 19.5k articles about cities around the world. After the step of filtering out problematic samples with incorrect sentence segmentation detected by mismatched counts between sentences and labels, the resulted dataset covers 21,376 articles with the highest-level section marks as ground-truth segment boundaries. We follow the setting in \citet{arnold-etal-2019-sector} by splitting the dataset into 70\% training, 10\% validation and 20\% test data.


\subsubsection{Domain Transfer Datasets}
To better evaluate models' robustness in cases where a domain-shift is present (called ``domain transfer segment inference''), we apply the topic segmenters trained on Wiki-Section to four small corpora heavily deviating from the training corpus (see Table~\ref{tab:stats_transfer} for more details): \\
\vspace{-2ex} \\
\textbf{Wiki-50 \cite{koshorek-etal-2018-text}} consists of 50 Wikipedia articles randomly sampled from the latest English Wikipedia dump. There is no overlap between this dataset and Wiki-Section. \\
\vspace{-2ex} \\
\textbf{Cities \cite{chen-etal-2009-global}} consists of 100 Wikipedia articles about cities. There is no overlap between this dataset and Wiki-Section, even the theme of this dataset is close to the portion of city articles in Wiki-Section. \\
\vspace{-2ex} \\
\textbf{Elements \cite{chen-etal-2009-global}} consists of 118 Wikipedia articles on chemical elements. \\
\vspace{-2ex} \\
\textbf{Clinical \cite{malioutov-barzilay-2006-minimum}} consists of 227 chapters in a clinical book. The subsection marks within each chapter are deemed as ground-truth segment boundaries.

\begin{table}
\centering
\scalebox{0.93}{
\begin{tabular}{l | @{\space\space\space} c@{\space\space\space\space} c@{\space\space\space\space} c@{\space\space\space}}

\specialrule{.1em}{.05em}{.05em}

\textbf{Dataset} & \# of doc & \# sent/seg & \# seg/doc \\
\hline
 WIKI-50 & 50 & 13.6 & 3.5 \\
 Cities & 100 & 5.2 & 12.2 \\
 Elements & 118 & 3.3 & 6.8 \\
 Clinical & 227 & 28.0 & 5.0 \\
\specialrule{.1em}{.05em}{.05em}
\end{tabular}
}
\caption{\label{tab:stats_transfer} Statistics of the datasets used in domain transfer experiments.}
\end{table}

\subsection{Experimental Design}
\paragraph{Baselines:} 
We directly compare our proposed discourse-aware topic segmentation model (called \textbf{Basic Model + Discourse}) with the following unsupervised and supervised baselines: \\
\vspace{-2ex} \\
\textbf{- BayesSeg} \cite{eisenstein-barzilay-2008-bayesian}: This unsupervised method makes segmentation prediction by situating the lexical cohesion of text in a Bayesian framework. A text span produced by a distinct lexical distribution is recognized as a coherent topic segment.\\
\vspace{-2ex} \\
\textbf{- GraphSeg} \cite{glavas-etal-2016-unsupervised}: This unsupervised method derives semantically coherent
segments through reasoning on a semantic relatedness graph construed from greedy lemma alignment.\\
\vspace{-2ex} \\
\textbf{- TextSeg} \cite{koshorek-etal-2018-text}: This supervised neural topic segmenter adopts a hierarchical neural sequence labeling framework with BiLSTM as the main architecture of each layer. The basic model used in our paper (described in section~\ref{sec:basic_model}) is an effective extension of this approach. \\
\vspace{-2ex} \\
\textbf{- Sector} \cite{arnold-etal-2019-sector}: This is a supervised neural topic segmenter extended from \textit{TextSeg} by adding an auxiliary layer for sentence topic label prediction. The learned intermediate topic embeddings for sentences are directly utilized for segment boundary inference.\\
\vspace{-2ex} \\
\textbf{- Transformer} \cite{glavas-2020-two}: This is a supervised neural topic segmenter consisting of two hierarchically connected Transformer networks for sentence encoding and sentence contextualization respectively.\\
\vspace{-2ex} \\
\textbf{- Basic Model + Context} \cite{xing-etal-2020-improving}: This is a top-performing neural topic segmenter which shares the same basic architecture with our proposal. The approach improves the \textbf{context modeling} capacity of the plain basic model by adding an auxiliary coherence prediction module and restricted self-attention. 

\paragraph{Evaluation Metrics:} We use the $P_k$ error score\footnote{We also considered \textit{windiff} \cite{pevzner-hearst-2002-critique} as another evaluation metric. Since it was highly correlated with $P_k$, we omit it and only present performance by $P_k$ to better compare with results reported in previous works.} \cite{Beeferman1999} for our intra-domain and domain transfer segment inference evaluations. The metric thereby simply measures the probability that a pair of sentences located at two ends of a $k$-sized sliding window in a document are incorrectly identified as belonging to the same segment or not. $k$ is determined as half of the average true segment size of the document. Since it is a penalty metric, lower values indicates better performance. We further quantitatively analyze models' efficiency according to two aspects: Model size and model speed, evaluating the count of learnable parameters and batches/documents processed per second during training/inference, besides $P_k$ measurement.

\paragraph{Implementation Details:}
For the hierarchical topic segmenter (our basic model), we adopt the default setting in \citet{xing-etal-2020-improving}, with GoogleNews word2vec ($d = 300$) as initial word embeddings and the contextualized representation of special token \texttt{[CLS]} ($d = 768$) from \texttt{bert-base-uncased} as initial sentence embeddings. All BiLSTM layers have the hidden state size = 256. For the discourse graph model component, the number of GAT layers is set to 2 through validation and the number of heads is set to 4 as in \cite{velickovic2018graph}. The input and output dimensions of each layer = 256. Training uses Adam with $lr = 1e^{-3}$ and batch size = 8. Early stopping is applied within 10 epoches of model training and the boundary prediction threshold $\tau$ is tuned over the validation set of each corpus we use for intra-domain model evaluation.

\subsection{Intra-Domain Segment Inference}
\label{sec:intra_domain}

\begin{table}
\centering
\scalebox{0.91}{
\begin{tabular}{l | c c c | c}

\specialrule{.08em}{.05em}{.05em}
\rowcolor{Gray}
\textbf{Dataset} & \textbf{Choi} & \textbf{Rules} & \textbf{Section} & \textbf{RSTDT} \\
 \hline
 Random & 49.4\textsuperscript{\space} & 50.6\textsuperscript{\space} & 51.3\textsuperscript{\space} & \cellcolor{green!15}40.5\textsuperscript{\space} \\
\hline
 BayesSeg & 20.8\textsuperscript{\space} & 41.5\textsuperscript{\space} & 39.5\textsuperscript{\space} & \cellcolor{green!15}37.5\textsuperscript{\space} \\
 GraphSeg & 6.6\textsuperscript{\space} & 39.3\textsuperscript{\space} & 44.9\textsuperscript{\space} & \cellcolor{green!15}58.7\textsuperscript{\space} \\
 TextSeg & 1.0\textsuperscript{\space} & 7.7\textsuperscript{\space} & 12.6\textsuperscript{\space} & \cellcolor{green!15}26.9\textsuperscript{\space} \\
 Sector  & --\textsuperscript{\space} & --\textsuperscript{\space} & 12.7\textsuperscript{\space} & \cellcolor{green!15}--\textsuperscript{\space}\\
 Transformer & 4.8\textsuperscript{\space} & 9.6\textsuperscript{\space} & 13.6\textsuperscript{\space} & \cellcolor{green!15}--\textsuperscript{\space}\\
 \hline
 Basic Model& 0.81\textsuperscript{\space} & 7.0\textsuperscript{\space} & 11.3\textsuperscript{\space} & \cellcolor{green!15}26.9\textsuperscript{\space} \\
 +Context & \textbf{0.54}\textsuperscript{\space} & \textbf{5.8}\textsuperscript{\space} & \textbf{9.7}\textsuperscript{\space} & \cellcolor{green!15}\underline{25.4}\textsuperscript{\space}\\
 \cellcolor{blue!10}+Discourse & \cellcolor{blue!10}\underline{0.59}\textsuperscript{\space} & \cellcolor{blue!10}\underline{6.1}\textsuperscript{\space} & \cellcolor{blue!10}\underline{10.2}\textsuperscript{\space} & \cellcolor{blue!10}\textbf{24.8}\textsuperscript{\space}\\
\specialrule{.08em}{.05em}{.05em}
\end{tabular}
}
\caption{\label{tab:res_intra_domain} $P_k$ ($\downarrow$) error score on three corpora for intra-domain experiment. Results in \textbf{bold} and \underline{underlined} indicates the best and second best performance across all comparisons. The row in \textcolor{blue!45}{purple} is the results achieved by our proposal. The column in \textcolor{green!85}{green} is the results for RSTDT paragraph break prediction with gold discourse structures integrated.}
\end{table}

We report our results of the intra-domain segment inference on the Choi, Rules and Wiki-Section datasets in Table~\ref{tab:res_intra_domain}. For better performance comparison, the table is subdivided into three sub-tables: random baseline, previously proposed approaches and models build on top of the basic model we use. We observe that the basic model without any additinal components already outperforms alternative supervised and unsupervised segmenters. With the above-sentence discourse dependency information injected, as proposed in this paper, the method (named +Discourse) further improves the performance by a notable margin across all three corpora. We further find that our proposed approach does not achieve superior performances compared to the basic model enhanced with the context modeling strategy (+Context) in \citet{xing-etal-2020-improving}. We believe that
a possible explanation for this under-performance could be the upstream parsing error of the discourse dependency parser applied out-of-domain, oftentimes severly impairing the parsing performance \cite{huber-carenini-2019-predicting}. Therefore, we conduct an additional experiment on RST-DT due to the availability of gold discourse structures annotated by human for this corpus. With no human-annotated topic segment boundaries at hand, we use paragraph breaks contained in RST-DT articles as the ground-truth for training and testing of topic segmentation models. Our results in Table~\ref{tab:res_intra_domain} show that the quality of discourse structure is positively correlated with enlarged improvements achieved by our proposal. In this case, the upper bound achieved by integrating gold discourse structures can even outperform the basic model enhanced by context modeling (+Context).

\subsection{Domain Transfer Segment Inference}
\label{sec:domain_transfer}

\begin{table}
\centering
\scalebox{0.825}{
\begin{tabular}{l | c c c c}

\specialrule{.1em}{.05em}{.05em}
\rowcolor{Gray}
\textbf{Dataset} & \multicolumn{1}{c}{\textbf{Wiki-50}} & \multicolumn{1}{c}{\textbf{Cities}} & \multicolumn{1}{c}{\textbf{Elements}} & \multicolumn{1}{c}{\textbf{Clinical}} \\
 \hline
 Random & 52.7\textsuperscript{\space} & 47.1\textsuperscript{\space} & 50.1\textsuperscript{\space} & 44.1\textsuperscript{\space} \\
\hline
 BayesSeg & 49.2\textsuperscript{\space} & 36.2\textsuperscript{\space} & \textbf{35.6}\textsuperscript{\space} & 57.2\textsuperscript{\space} \\
 GraphSeg & 63.6\textsuperscript{\space} & 40.0\textsuperscript{\space} & 49.1\textsuperscript{\space} & 64.6\textsuperscript{\space} \\
 TextSeg & 28.5\textsuperscript{\space} & 19.8\textsuperscript{\space} & 43.9\textsuperscript{\space} & 36.6\textsuperscript{\space} \\
 Sector  & 28.6\textsuperscript{\space} & 33.4\textsuperscript{\space} & 42.8\textsuperscript{\space} & 36.9\textsuperscript{\space}\\
 Transformer & 29.3\textsuperscript{\space} & 20.2\textsuperscript{\space} & 45.2\textsuperscript{\space} & 35.6\textsuperscript{\space}\\
 \hline
 Basic Model& 28.7\textsuperscript{\space} & 17.9\textsuperscript{\space} & 43.5\textsuperscript{\space} & 33.8\textsuperscript{\space} \\
 +Context & \textbf{26.8}\textsuperscript{\space} & \textbf{16.1}\textsuperscript{\space} & \underline{39.4}\textsuperscript{\space} & \textbf{30.5}\textsuperscript{\space}\\
 \cellcolor{blue!10}+Discourse & \cellcolor{blue!10}\textbf{26.8}\textsuperscript{\space} & \cellcolor{blue!10}\underline{16.9}\textsuperscript{\space} & \cellcolor{blue!10}41.1\textsuperscript{\space} & \cellcolor{blue!10}\underline{31.8}\textsuperscript{\space}\\
\specialrule{.1em}{.05em}{.05em}
\end{tabular}
}
\caption{\label{tab:res_transfer} $P_k$ ($\downarrow$) error score on four test corpora for domain transfer experiment. Results in \textbf{bold} and \underline{underlined} indicates the best and second best performance across all comparisons. The row highlighted in \textcolor{blue!45}{purple} is the results achieved by our proposal.}
\end{table}

Table~\ref{tab:res_transfer} presents the performance of simple baselines, previously proposed models and our new approach on the domain transfer task. 
Similar to the intra-domain segment inference, the Basic Model+Context approach still achieves the best performance across all testing domains except Elements, in which the unsupervised BayesSeg performs superior. However, our +Discourse strategy still leads to improvement over the basic model, and achieves comparable performance to the best model (+Context) on Wiki-50 and Cities. We believe that it gives evidence that injecting discourse dependency structures has potential to enhance the generality of topic segmentation models. 

\subsection{Efficiency Analysis}

Table~\ref{tab:res_efficiency} compares the efficiency of the top two models, comparing our proposed approach (Basic Model+Discourse) against Basic Model+Context. The experiments for these systems were carried out on a Nvidia Telsa V100 16G GPU card. 
We observe that our strategy of injecting discourse dependency structures can improve model's performance on intra-domain and domain transfer setting, but with less increase of model size and loss of speed compared to +Context.
More specifically, adding our discourse graph modeling component on top of the basic model introduces 65\% more learnable parameters while the context modeling components in \citet{xing-etal-2020-improving} cause a 127\% parameter increasing. On the other hand, discourse graph modeling slightly slows down the speed of model training and inference by 21\% and 7.7\% respectively, while making more complex context modeling significantly slows down the speed by 78\% and 46\%. Together with the previous results about model's effectiveness, we can see that our proposed system would be a better option in practical settings where efficiency is critical.

Additionally, we conduct the same set of experiments for the model with both context modeling module and our proposed discourse structure integration (Basic Model+Context+Discourse). The performance of this model always falls in between +Context and +Discourse individually, but with the worst efficiency measured by model size and speed.

\begin{table}
\centering
\scalebox{0.87}{
\begin{tabular}{l | c c c}

\specialrule{.1em}{.05em}{.05em}
\rowcolor{Gray}
 & \multicolumn{1}{c}{\textbf{\# Params $\downarrow$}} & \multicolumn{1}{c}{\textbf{T-Speed $\uparrow$}} & \multicolumn{1}{c}{\textbf{I-Speed $\uparrow$}} \\
 \hline
 Basic Model & 4.82M\textsuperscript{\space} & 6.90\textsuperscript{\space} & 35.58\textsuperscript{\space} \\
 \hline
 +Context & 10.93M\textsuperscript{\space} & 1.49\textsuperscript{\space} & 19.23\textsuperscript{\space} \\
 +Discourse & \textbf{7.97M}\textsuperscript{\space} & \textbf{5.44}\textsuperscript{\space} & \textbf{32.85}\textsuperscript{\space} \\
\specialrule{.1em}{.05em}{.05em}
\end{tabular}
}
\caption{\label{tab:res_efficiency} The efficiency comparison between our proposal and the method proposed in \citet{xing-etal-2020-improving} on the Wiki-Section corpus. These two models share the same basic segmentation framework. \textbf{T-Speed} refers the training speed as number of batches processed per second during training stage. \textbf{I-Speed} refers the inference speed as number of documents processed per second during inference stage.}
\end{table}

\section{Conclusion and Future Work}

In this paper, we present a neural topic segmentation model with injection of above-sentence discourse dependency structures inferred from a state-of-the-art discourse dependency parser. Different from previously proposed methods, our segmenter leverages the discourse signal by encoding the topical consistency between sentences from a more global and interpretable point of view. Experiments on multiple settings (intra-domain, domain transfer and efficiency comparison) show that our system achieves comparable performance to one of the current top-performing topic segmenters, with much less model size increase and speed degradation.

In the near future, we plan to investigate the synergy between topic segmentation and discourse parsing more comprehensively, by incorporating the type of inter-sentential rhetorical relations and analyzing whether and how this discourse knowledge can enhance supervised topic segmentation frameworks. In the long run, we intend to explore the possibility for discourse parsing to benefit segment topic labeling, which is another important task usually coupled together with topic segmentation to provide the coarse-grained structural information for documents. Particularly, we believe discourse parsing can potentially enhance the step of key phrase extraction in segment topic labeling due to the significant improvement it brings to the related task of name entity recognition (NER) \cite{jie-lu-2019-dependency}.

\section*{Acknowledgments}
We thank the anonymous reviewers and the UBC-NLP group for their insightful comments and suggestions. This research was supported by the Language \& Speech Innovation Lab of Cloud BU, Huawei Technologies Co., Ltd.

\bibliography{anthology,acl2020}
\bibliographystyle{acl_natbib}

\end{document}